\documentclass{article}
\pdfoutput=1

\PassOptionsToPackage{numbers, compress}{natbib}

\usepackage[preprint]{neurips_2023}

\usepackage[utf8]{inputenc} 
\usepackage[T1]{fontenc}    
\usepackage[colorlinks=true,citecolor=red]{hyperref} 
\usepackage{url}            
\usepackage{booktabs}       
\usepackage{amsfonts}       
\usepackage{nicefrac}       
\usepackage{microtype}      
\usepackage{xcolor}         

\usepackage{color}
\usepackage{arydshln}
\usepackage{graphicx}
\usepackage{amsmath,bm}
\usepackage{amssymb}
\usepackage{array}
\usepackage{comment}
\usepackage{booktabs}
\usepackage{subfigure}
\usepackage{caption}
\usepackage{colortbl} 
\usepackage{multirow}
\usepackage{makecell}
\usepackage{wrapfig}

\usepackage[ruled,vlined]{algorithm2e}
\usepackage[normalem]{ulem}
\usepackage{arydshln} 
\usepackage{enumitem}

\usepackage{pifont}

\usepackage{tabularray}
\usepackage{paralist}
\usepackage{makecell}
\usepackage{colortbl}
\usepackage{etoolbox}

\usepackage{rotating}

\definecolor{myblue}{RGB}{52,218,247}
\definecolor{myred}{RGB}{255,90,90}
\definecolor{mypink}{RGB}{239,43,159}
\definecolor{myupdate}{RGB}{254,243,222}
\definecolor{myfrozen}{RGB}{237,255,255}
\definecolor{ired}{RGB}{229,72,72}
\definecolor{igreen}{RGB}{80,219,144}
\definecolor{ired}{RGB}{247,142,142}

\newcommand{\etal}{{et al}.\@ }

\newcommand{\eg}{\textit{e}.\textit{g}., }

\title{Towards Complex-query Referring Image Segmentation: A Novel Benchmark}

\author{%
Wei Ji \quad Li Li \quad Hao Fei \quad Xiangyan Liu \quad Xun Yang \quad Juncheng Li \quad Roger Zimmermann \\
National University of Singapore
}

\begin{document}

\maketitle

\begin{abstract}
Referring Image Understanding (RIS) has been extensively studied over the past decade, leading to the development of advanced algorithms. 
However, there has been a lack of research investigating how existing algorithms should be benchmarked with complex language queries, which include more informative descriptions of surrounding objects and backgrounds (\eg \textit{"the black car."} vs. \textit{"the black car is parking on the road and beside the bus."}). 
Given the significant improvement in the semantic understanding capability of large pre-trained models, it is crucial to take a step further in RIS by incorporating complex language that resembles real-world applications.
To close this gap, building upon the existing RefCOCO and Visual Genome datasets, we propose a new RIS benchmark with complex queries, namely \textbf{RIS-CQ}. 
The RIS-CQ dataset is of high quality and large scale, which challenges the existing RIS with enriched, specific and informative queries, and enables a more realistic scenario of RIS research.
Besides, we present a nichetargeting method to better task the RIS-CQ, called dual-modality graph alignment model (\textbf{\textsc{DuMoGa}}), which outperforms a series of RIS methods. 
\end{abstract}

\section{Introduction}
\label{Introduction}

\begin{figure}[t]
\centering  
\includegraphics[width=0.9\linewidth]{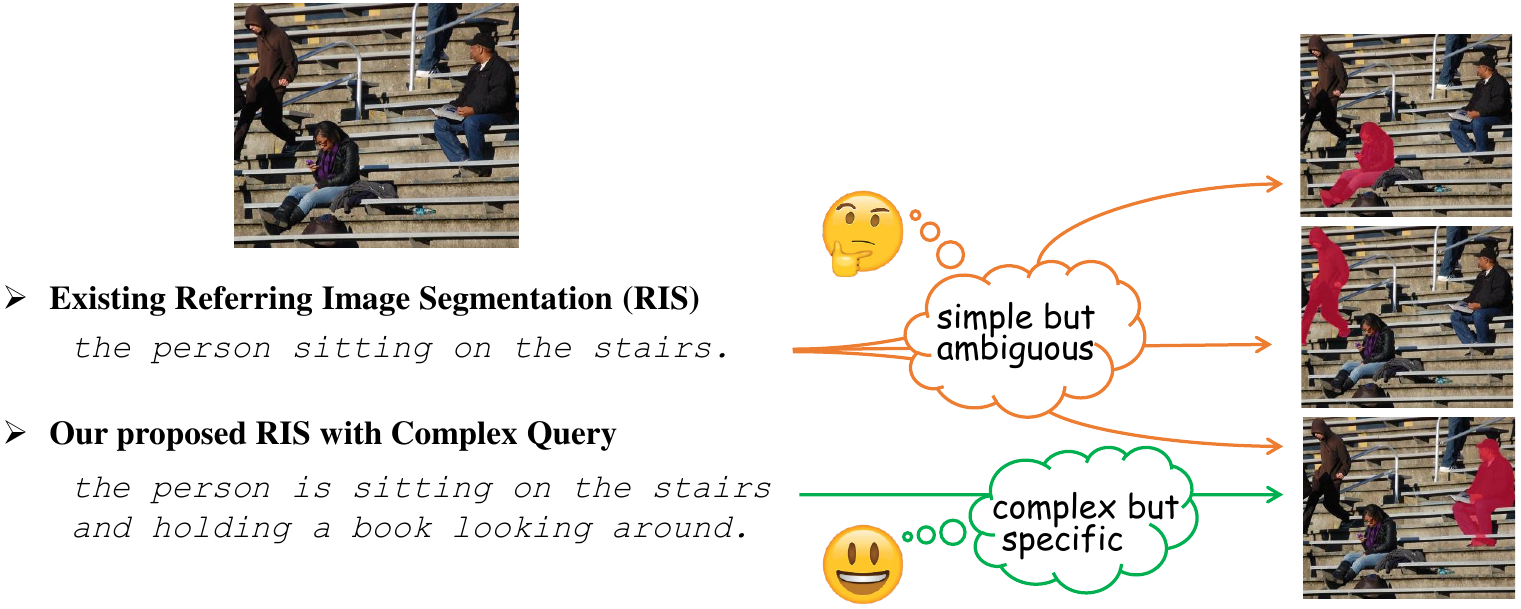}

\caption{
The comparison between the existing Referring Image Segmentation (RIS) and the complex-query RIS proposed in this work.
} 
\label{fig:intro}

\end{figure}

Correctly comprehending the subtle alignment (i.e., grounding) between language and vision has long been the pivotal research in the intersecting communities of computer vision and language processing \cite{yang2020improving,yang2019fast,zhao2022word2pix}.
Among a range of vision-language grounding topics, Referring Image Segmentation (RIS) has been proposed with the aim to ground a given language query onto a specific region of an image, i.e., typically represented by a segmentation map, as exemplified in Figure \ref{fig:intro}. 
By precisely bridging the semantics between the images and texts, RIS plays a crucial role in various downstream cross-modal applications, including image editing \cite{chen2018language,li2020manigan}, and language-based human-robot interaction\cite{wang2019reinforced,zhang2020multimodal}.


Although attracting increasing research attention, the existing RIS benchmarks, unfortunately, can be subject to the \textbf{simplicity and naivety of the language queries}.
Specifically, via data statistics of the current popular RIS datasets, including RefCOCO and RefCOCO+ \cite{yu2016modeling,mao2016generation}, we find that \emph{i)} 85.3\% text queries have short forms (i.e., with length $\leq$ 5 words), and \emph{ii)} 83.8\% queries involve only one or two visual objects.
We note that such characteristics would inevitably lead to the key pitfalls that hamper the utility and applicability of RIS task.

\setdefaultleftmargin{2.0em}{2em}{2.0em}{1.0em}{1em}{1em}
\begin{compactitem}
    \item On the one hand, caused by the inequality between vision and language, i.e., language is abstract and succinct while vision always entails richer details, 
    simple short language queries with shallow object descriptions usually trigger ambiguity for the visual coreference.
    The sample in Figure \ref{fig:intro} shows the case.
    More severely, in the existing RIS datasets, a significant majority of query sentences suffer from such ambiguity.
    Intuitively, RIS models, being trained on such overly simplistic data, would largely fail to disambiguate the image referring and lead to suboptimal performance.

    




    \item On the other hand, in real-world applications, individuals are more likely to input detailed textual descriptions in complex forms, so as to accurately locate the desired objects.
    Although the recent RIS methods \cite{yang2022lavt,yu2018mattnet,feng2021encoder} achieve satisfactory performance on the in-house testing set, they can still struggle when confronted with complex and informative queries.
    Our preliminary experiment indicates that even the current top-performing RIS model, LAVT\cite{yang2022lavt}, drops dramatically (74.46 vs. 10.84 in mIoU on two RIS datasets) when testing on the complex query.
    In other words, being trained on existing RIS datasets can lead to the failure of generalizing to the input queries from realistic users in the wild.




    
\end{compactitem}




The above observations imperatively motivate the exploration of \emph{Referring Image Segmentation with Complex Queries (RIS-CQ)}.
In this work, we propose a novel benchmark for RIS under such complex scenario.
Specifically, we construct a RIS-CQ dataset (cf. \S\ref{Constructing Complex Query for Referring Image Segmentation}), which is of high quality and large scale.
Technically, we first extract the salient objects with their pertaining relations (spatial, action, \emph{etc.}) from an image, and then generate semantically detailed and meaningful descriptions for the referring objects towards their surrounding objects, which serve as the complex-form queries.
Notedly, the recent large language models (LLMs), e.g., ChatGPT \cite{lund2023chatting} have revealed the great capability of human-level understanding of language semantics.
Thus we take advantage of the LLM to assist to generate large amounts of complex queries while without sacrificing the labeling quality.
Finally, we obtain a novel RIS-CQ dataset with 118,287 images and 13.18 words on average for each query.

The key to accurate RIS recognition in our scenario essentially lies in the deep understanding of the underlying semantics of different modalities, because of the intrinsic complex form of textual query and the sophisticated visual content.
To this end, we propose a novel \textbf{du}al-\textbf{mo}dality \textbf{g}raph \textbf{a}lignment (dubbed \textbf{\textsc{DuMoGa}}) model (cf. \S\ref{Proposed Method}) to benchmark the RIS-CQ task, where the input sentence and image are represented with the semantic dependency graph \cite{peters-etal-2018-deep} and semantic scene graph \cite{tang2019learning}, respectively.
Meanwhile, the semantics of the two modalities should be sufficiently interacted for deep comprehension of the inputs.
Correspondingly, two levels of cross-modal alignment learning are carried out in \textsc{DuMoGa} to capture the intrinsic correspondence between the input text and vision, including the structural alignment between two semantic graphs and the feature alignment between two semantic representations.
On the RIS-CQ datasets our \textsc{DuMoGa} achieves 24.4 in mIoU, outperforming the current state-of-the-art (SoTA) RIS method by more than 200\%.

To sum up, this work contributes to the following three aspects.
\textbf{(1)} We construct a novel benchmark dataset, RIS-CQ, which challenges the existing RIS with complex queries, and enables a more realistic scenario of RIS research.
\textbf{(2)} We present a strong-performing system (\textsc{DuMoGa}) to model the task, which brings a new SoTA results on the RIS-CQ dataset.
\textbf{(3)} Series of in-depth analyses are shown based on our dataset and the systems, where some important and interesting findings are presented and concluded to shed light on the future exploration of this topic.
All our data and resource will be made open later to facilitate the follow-up research.

\section{Constructing Complex Query for Referring Image Segmentation}
\label{Constructing Complex Query for Referring Image Segmentation}

\subsection{Problem Definition}

Given an image $\mathcal{I}$ and a set of textual queries (i.e., long-form sentences with semantically complex expressions) $\mathcal{Q} = \{ \mathbf{p}_i \}_{i = 1}^M$, 
RIS-CQ aims to predict a set of 
segmentation masks $\mathcal{S} = \{ \mathbf{s}_i \}_{i = 1}^M$, where each 
$\mathbf{s}_i$ correspond to each query $\mathbf{p}_i$ that localizes it in the image.
Note that $M$ is the number of referring expressions for a given image $\mathcal{I}$, which is set as $M \in [1,10]$.


\subsection{Dataset Construction}

In this section, we elaborate on how to elicit complex and contextualized queries from large language models~(\textsl{i.e.}, ChatGPT) by leveraging diverse structured semantics in images (\textsl{e.g.}, inter-object relations). First, we extract holistic semantic relations for each image and then select more significant objects that involve diverse interactions with other objects as the candidate objects. Second, we utilize specially tailored prompts to guide ChatGPT in generating complex queries for each candidate object based on rich visual context. Finally, we manually filter out or revise problematic queries (\textsl{e.g.}, ambiguous references), to ensure the annotation quality.

\paragraph{Step-1: Relation extraction.}
To begin with, we utilize an off-the-shelf scene graph generation model, VC-Tree, to directly extract the objects present in the images along with their relationships. 
These relationships are stored in the form of triplets, such as <person, cup, holding>, which indicates that the person is holding the cup. Additionally, during this step, we apply a simple filter to exclude objects that have fewer than \textbf{two} relationships with other objects. We believe that generating queries to describe these objects would be relatively straightforward, increasing the probability of ambiguous references or incorrect references.

\paragraph{Step-2: Complex query generation.}
In this section, we devise a tailored set of prompts to efficiently leverage the in-context learning capabilities of a large language model. For each object and its corresponding relation triplets, we generate a descriptive text query using the model. For instance, given a set of triplets associated with the target object \textit{person}, {<person, cup, holding>, <person, wall, leaning on>, <table, person, next to>}, ChatGPT can provide us with the output "the person is next to the table, holding the cup and leaning on the wall."

Specifically, we utilize the gpt-3.5-turbo model as our large language model, with the API interface provided by OpenAI. To ensure stable outputs from the model, we set the decoding temperature to 0. The relevant prompts used are detailed in the appendix for reference. An additional noteworthy aspect is that the presence of the large language model entirely liberates us from the manual labor of annotating images for generating language queries. This significantly reduces our workforce costs while ensuring the quality of queries. Moreover, it facilitates the expansion of our dataset to the magnitude of 100k.

\begin{figure}[t]
  \centering  \includegraphics[width=0.98\linewidth]{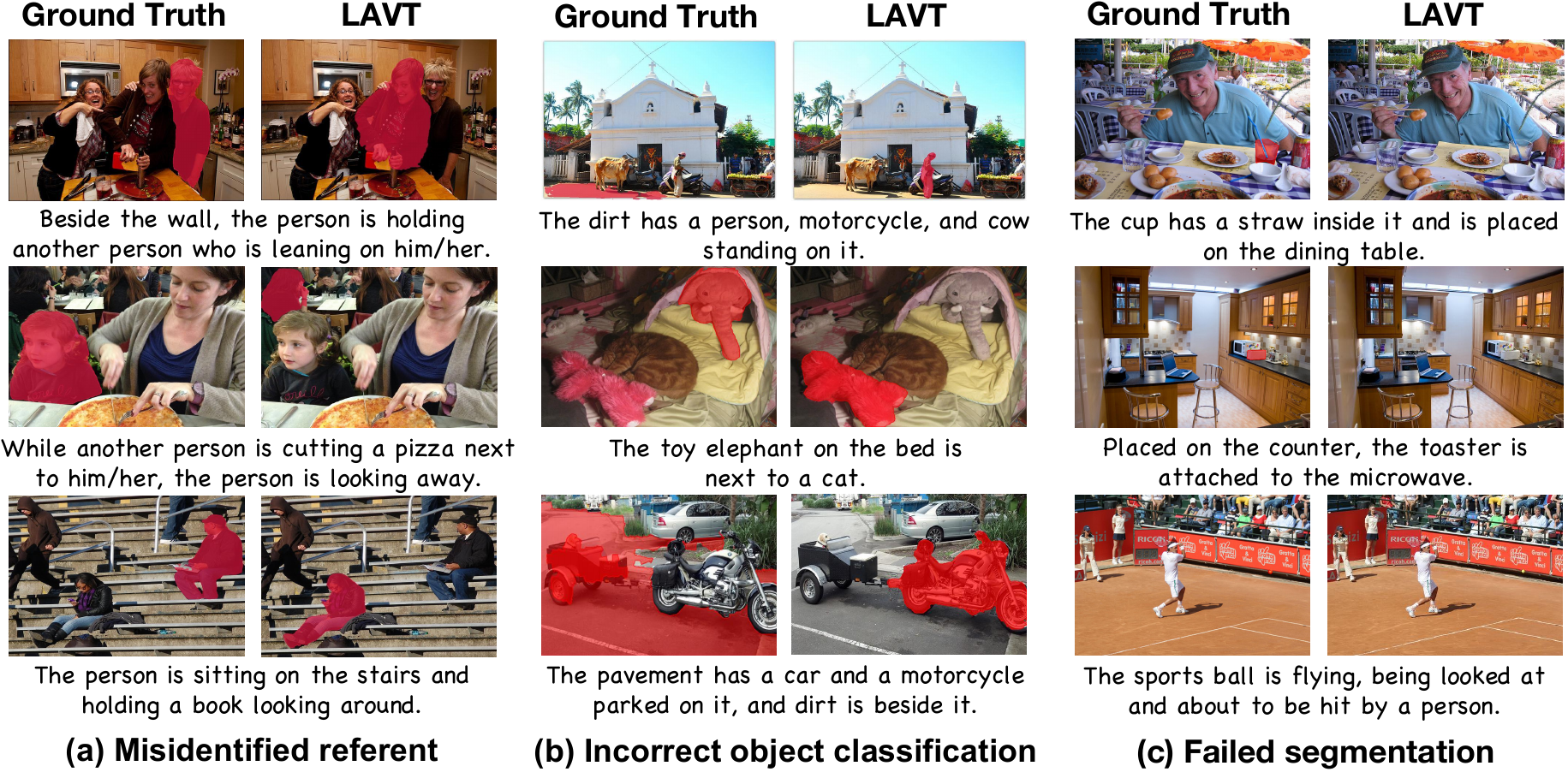}

  \caption{Failure cases of Complex-query Referring Image Segmentation.} 
  \label{fig:failure}

\end{figure}

\paragraph{Step-3: Post-processing.}
Finally, we manually filter out or revise queries with ambiguous references to ensure a precise one-to-one correspondence between the queries and objects within the image.
After multiple iterations in constructing prompts, we have successfully developed prompts that yield high-quality queries generated by ChatGPT. 
However, it is important to note that since ChatGPT receives input solely from the relation triplets generated by the scene graph model, it lacks certain contextual information from the image. 
As a result, there may be instances where the generated queries exhibit ambiguous references to the objects being described. 
Unfortunately, these queries cannot be filtered out automatically in our automated pipeline, necessitating manual intervention for their removal.







\subsection{Dataset Statistics}

\begin{figure*}[t]
    \begin{minipage}[t]{0.98\linewidth}
      \centering
      \includegraphics[width=\linewidth]{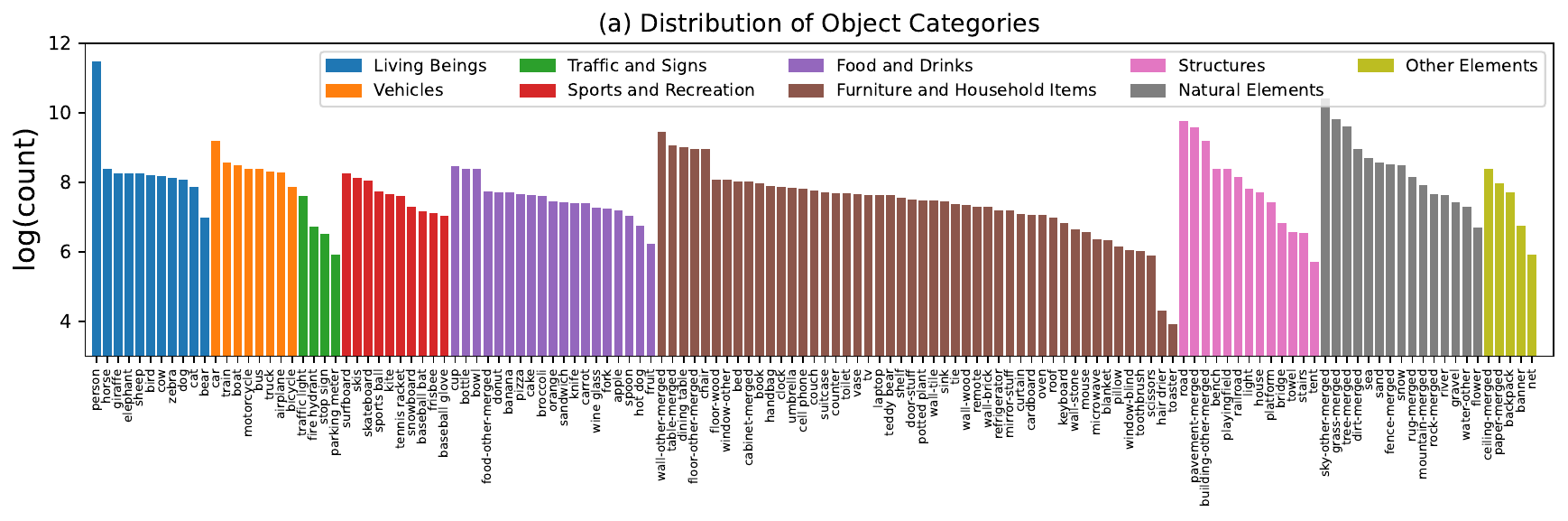}
    \end{minipage}
    \begin{minipage}[t]{\linewidth}
      \centering
      \includegraphics[width=\linewidth]{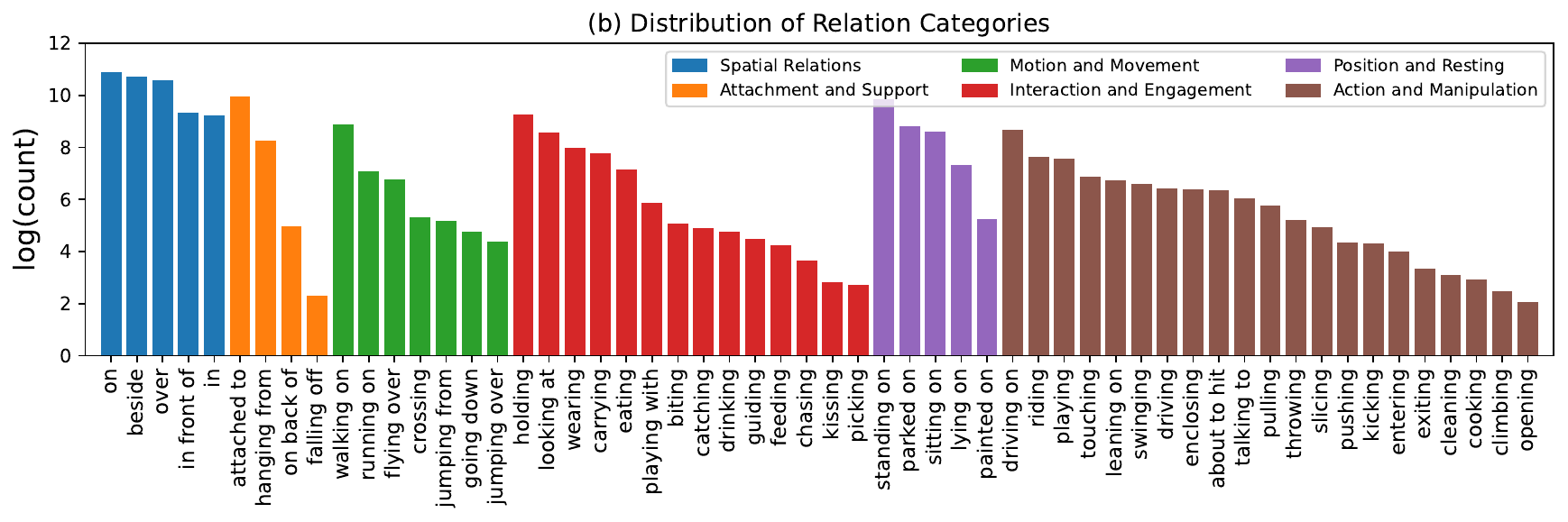}
    \end{minipage}

  \caption{The distribution of object and relation categories, organized based on the parent classes.
  Best viewed by zooming in.
  }
  \label{fig:stat}

\end{figure*}

Figure \ref{fig:stat} presents the statistical analysis of objects and predicates in the RIS-CQ dataset, which consists of a total of 133 object classes and 56 relation classes. The language queries generated are solely based on the objects depicted in Figure \ref{fig:stat} along with their corresponding relations. The objects and relations are categorized into 9 groups and 6 groups respectively, with each group containing elements that exhibit certain correlations. The group names represent abstract summaries of the shared attributes among the elements within each group.

We analyze the failure cases of current models on RIS-CQ dataset. 
From Figure ~\ref{fig:failure}, we can summarize them into three types: misidentified referent, incorrect object classification, and failure segmentation.


\subsection{Dataset Comparison}

\noindent{\bf RefCOCO/ RefCOCO+/ RefCOCOg.}  RefCOCO~\cite{yu2016modeling}, RefCOCO+~\cite{yu2016modeling}, and RefCOCOg~\cite{mao2016generation} are three visual grounding datasets with images and referred objects selected from MS COCO~\cite{lin2014microsoft}. The referred objects are selected from the MS COCO object detection annotations and belong to 80 object classes. RefCOCO~\cite{yu2016modeling} has 19,994 images with 142,210 referring expressions for 50,000 object instances. RefCOCO+ has 19,992 images with 141,564 referring expressions for 49,856 object instances. RefCOCOg has 25,799 images with 95,010 referring expressions for 49,822 object instances. 



\paragraph{Visual Genome.}
Visual Genome (VG v1.4) \cite{krishna2017visual} contains $108,077$ images with $21$ relationships on average per image, which is split into $103,077$ training images and $5,000$ testing images.

\begin{table*}[!ht]
\fontsize{9}{11.5}\selectfont
\tabcolsep=0.14cm
    \centering
    \caption{Statistics of current Referring Image Understanding benchmarks.}
    \label{table:dataset}
    \begin{tabular}{cccccc}
    \hline
        \bf Dataset & \bf RefCOCO & \bf RefCOCO+ & \bf RefCOCOg & \bf Visual Genome & \bf RIS-CQ \\ \hline 
        \# Images & 19,994 & 19,992 & 26,711 & 108,077 & 118,287 \\
        \# Text query & 142,209 & 141,564 & 85,474 & 5.4M & 285,781 \\ 
        Avg. query length & 3.61 & 3.53 & 8.43 & 5 & 13.18 \\ 
        Avg. object / query & 1.76 & 1.67 & 3.03 & - & 3.58 \\ 
        Annotation methods & Manual & Manual & Manual & Manual & Auto + Manual\\ \hline
    \end{tabular}

\end{table*}



\paragraph{RIS-CQ.} is our proposed \ referring image segmentation benchmark, which targets the explanation of image contents. The image source for RIS-CQ is the union of VG \cite{krishna2017visual} and COCO \cite{lin2014microsoft} datasets. RIS-CQ challenges RIS models to understand the rich object interactions in daily activities. 
As shown in Table \ref{table:dataset}, we summarize the details of each dataset compared with our RIS-CQ dataset.

\section{Proposed Method}
\label{Proposed Method}

\begin{figure}[t]
  \centering  \includegraphics[width=1\textwidth]{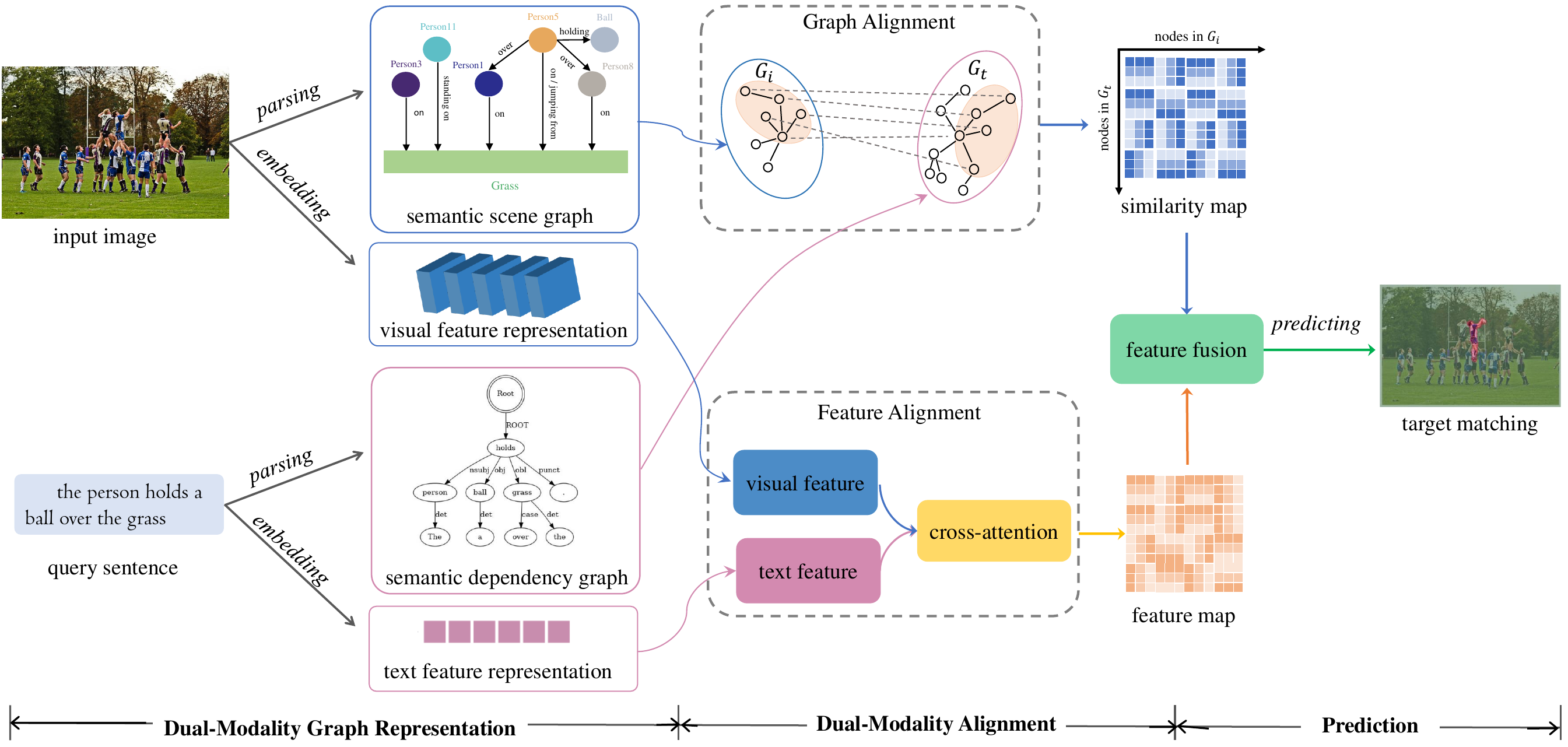}
  \caption{
  Our proposed \textsc{DuMoGa} framework, includes three procedures: Dual-modality Graph Representation, Dual-modality Graph Alignment, and Prediction.
  Best viewed by zooming in.
  } 

  \label{fig:method}
\end{figure}

Unlike classic dense prediction models with complex design in RIS (e.g. LAVT \cite{yang2022lavt}), which take expensive computational power to make inferences, the graph learning based architecture provides efficient training process and promising results. To address the numerous descriptions of surrounding objects and backgrounds within images, images are parsed into scene graphs, which are treated as fine-grained image representations to analyze the relations of these objects with a graph structure. On the other hand, following previous dependency-based RE methods \cite{guo-etal-2019-attention}, we parse the queries into syntax dependency trees, which can provide information for dual-modality graph alignment. As a result, we propose a graph learning-based method named \textsc{DuMoGa} to align the semantics in queries and the information in images, aiming at efficiently locating the target instances. The whole structure of our proposed \textsc{DuMoGa} is shown in Fig ~\ref{fig:method}.

\subsection{Dual-Modality Graph Representation}
\noindent{\textbf{Semantic Scene Graph Generation for Vision.}} We employ a classic SGG model (e.g. VCTree \cite{tang2019learning,li2023panoptic}) to parse images into scene graphs. 
The whole process can be formulated as:
\begin{equation}
    P_r\left(\mathcal{G}\ |\ \mathcal{I}\right)=P_r\left(\mathcal{S},\mathcal{O},\mathcal{R}\ |\ \mathcal{I}\right).
\end{equation}

An image $\mathcal{I} $ can be segmented into a set of masks $\mathcal{S} $. Each mask is associated with an object with class label $\mathcal{O}$. A set of relations $\mathcal{R}$ between objects are predicted. We construct the node set $ N_i=\left\{O_j|j\in\left[1,n\right]\right\} $ of the scene graph, $ O_i $ is the $ i_{th} $ object detected in the image. $ n $ represents the number of detected object, and we set the maximum number of detected object for each image to \textit{N}. For the edge set $ E_i=\left\{R_{a,b}=\left(I_a,I_b\right)|a\in\left[1,n\right],b\in\left[0,n-1\right]\right\} $, $ R_{a,b} $ denotes the $ a_{th} $ object related with the $ b_{th} $ object. As a result, the scene graph can be formulated as follows:
\setlength\abovedisplayskip{2pt}
\setlength\belowdisplayskip{2pt}
\begin{equation}
    G_i=\left(N_i,E_i\right).
\end{equation}

\noindent{\textbf{Syntax Dependency Graph for Text.}} We use the syntax dependency tree to explore the dependency between words in the sentence. Following \cite{peters-etal-2018-deep}, we use ELMo to obtain the dependency tree for the input text after which each word from the text is connected by its governor and obtains its related dependency triple. For the node set $ N_t=\left\{T_j|j\in\left[1,l\right]\right\} $ of the dependency tree, $ T_j $ denotes the $ j_{th} $ token in the sentence, and $ l $ represents the total length of the sentence. For the edge set $ E_t=\left\{G_{a,b}=\left(T_a^*,T_a\right)|a\in\left[1,l\right]\right\} $,$ T_a^* $ denotes the governor of the $ a_{th} $ token, and the graph representation for the sentence can be formulated as follows:
\setlength\abovedisplayskip{2pt}
\setlength\belowdisplayskip{2pt}
\begin{equation}
    G_t=\left(N_t,E_t\right).
\end{equation}

\subsection{Dual-modality Alignment}

To accurately locate the target instance using the query sentence, the gap between visual and semantics needs to be bridged. Thus, we propose the graph and feature alignment process to efficiently align visual and language domains.

\noindent{\textbf{Graph Alignment.}} We approximate the node embedding matrix $ \widetilde{P} $ by factorizing a similarity matrix of the node identities, and align nodes between the above two graphs by greedily matching the most similar embeddings from the other graph. 

Following \cite{zheng2021multi}, we combine $ N_i $ and $ N_t $, and count both in and out degrees of k-top neighbors for each node. Then we compute the similarity between every two nodes, getting a $ n \times n $ similarity matrix $\widetilde{P}$. After that we subsets $ P_1 $ and $ P_2 $ from $\widetilde{P}$. The $ P_1 $ and $ P_2 $ denote separate representations for nodes in $ G_i $ and $ G_t $.
\setlength\abovedisplayskip{2pt}
\setlength\belowdisplayskip{2pt}
\begin{equation}
    {\widetilde{P}}_1,{\widetilde{P}}_2=D\left(N\left(\widetilde{P}\right)\right),
\end{equation}
where $D$ denotes the dividing operation of $P$ by the number of nodes in $N_i$ and $N_t$ in order, and $N$ is normalization operation. When finishing graph structure alignment, we calculate the similarity between node \textit{i} from ${\widetilde{P}}_1$ and node \textit{j} from ${\widetilde{P}}_2$ using the formula below:

\begin{equation}
    a_{ij}=exp(-\left \| {\widetilde{P}}_1[i]-{\widetilde{P}}_2[j] \right \|_{2}^{2}  ).
\end{equation}

With the similarities between nodes, the two graphs are transformed into a similarity map $ \alpha $, which can be formulated as:
\setlength\abovedisplayskip{2pt}
\setlength\belowdisplayskip{2pt}
\begin{equation}
    \alpha=\left(a_{ij}\right)_{|V_1|\times|V_2|},
\end{equation}
where $ a_{ij} $ represents the structural similarity between the $i_{th}$ word of the input text and the $j_{th}$ object of the input image.

\noindent{\textbf{Feature Alignment.}} 
Though graph structure is efficient, the information within is not enough for the task. We take the advantage of visual features from images and word embeddings from sentences to promise a fine-grained searching space. Specifically, for each image $I$, we get its visual feature $F_i$ from backbone model (e.g. ResNet-50 \cite{he2016deep}). For each sentence $Q$, we get its semantic embedding $F_l$ using BERT \cite{devlin-etal-2019-bert}. We treat $F_i$ as the query while $F_l$ is treated as the key and value, completing the attention process below:
\setlength\abovedisplayskip{2pt}
\setlength\belowdisplayskip{2pt}
\begin{equation}
    R^{a}=\text{Softmax}\left(\frac{qk^T}{\sqrt d}\right)v,
\end{equation}
where d denotes the dimension of $F_l$.

\noindent{\textbf{Feature Fusion and Prediction.}} 
After obtaining the results from graph alignment and feature alignment processes, we fuse them to make the final prediction.
\setlength\abovedisplayskip{2pt}
\setlength\belowdisplayskip{2pt}
\begin{equation}
    R^{f}=MLP\left([R^{a} ; \alpha F_l]\right), \quad
    s = j^{*} = \mathop{\mathrm{Argmax}}_{1\leq j\leq N}(R^{f}_{1},...R^{f}_{j},...,R_{N}^f)    \,,
\end{equation}
Where $[;]$ denotes concatenation, and we use the MLPs to project the feature to \textit{N} dimension. 
$s$ denotes the output of the referring objects from the input with the largest probability.



\subsection{Training Loss}
For the \textit{N} detected objects in the image, we match their masks with the ground truth mask and obtain the $g_{th}$ object that share the most intersection over union with the ground truth. To maximize the probability on the $g_{th}$ dimension of $R^f$, we employ a cross-entropy loss function below:

\begin{equation}
    L= {\textstyle \sum_{i=1}^{N}} [-K^ilog(R_{i}^{f} )-(1-K^i)log(1-R_{i}^{f})].
\end{equation}

\textit{K} represents a one-hot vector with all zeros except for the $ g_{th} $ dimension, and $ K^i $ denotes the $ i_{th} $ dimension of \textit{K}. As a result, the probability on the $g_{th}$ dimension of $R^f$ is maximized.

\section{Experiment}

\subsection{Experimental Settings}
We employ a VCTree \cite{tang2019learning} with ResNet-50 \cite{he2016deep} as its backbone for scene graph generation and visual feature extraction. The maximum detected object in one image is set to 10. Specifically, the feature dimension for each of the detected object is 1024. For the query sentence, we append the [CLS] and [SEP] tokens to the beginning and end of the sentence respectively. Then a pre-trained BERT is used to generate sentence embedding and the dimension is set to 768. For our \textsc{DuMoGa} model, it is trained with Adamw optimizer, where we set the base learning rate at 2e-5 and the batch size at 64. 

\begin{table}[!t]
\centering
\fontsize{9}{12}\selectfont
 \setlength{\tabcolsep}{1.8mm}

\caption{Comparison with state-of-the-art methods in terms of overall IoU on three benchmark datasets. \textit{GA} represents Graph Alignment, \textit{FA} represents Feature Alignment, and \textit{Full} represents Graph Alignment + Feature Alignment.}
\label{tab:overall_performance}

\begin{tabular}{lccccccc}
\toprule
\multirow{2}{*}{\bf Method} & \multirow{2}{*}{\bf Backbone Model} & \multicolumn{6}{c}{\bf Refer Image Segmentation}                     \\ \cline{3-8} 
                        &                                 & \multicolumn{1}{c}{mIoU} & P@0.3 & P@0.4 & P@0.5 & P@0.6 & P@0.7 \\ 
\hline
MAttNet \cite{yu2018mattnet}                & ResNet-101                    & \multicolumn{1}{c}{8.00} & 9.61  & 7.90  & 6.15  & 5.51  & 4.58  \\
VPD \cite{zhao2023unleashing}                   & U-Net                      & \multicolumn{1}{c}{24.0} & 29.5  & 27.5  & 23.8  & 21.5   & \textbf{19.3}   \\ 
LAVT \cite{ding2021vision}                   & SWIN-B                        & \multicolumn{1}{c}{21.2} & 26.6  & 21.6  & 17.1  & 13.7   & 10.9   \\ 
UNINEXT \cite{UNINEXT}                   & ResNet-50                      & \multicolumn{1}{c}{19.8} & 22.3  & 21.8  & 21.0  & 19.9   & 19.2   \\ 
\hline
\textsc{DuMoGa} (\textit{GA})             & ResNet-50                       & \multicolumn{1}{c}{15.0} & 19.5  & 18.3  & 16.7  & 14.4  & 11.7  \\ 
\textsc{DuMoGa} (\textit{FA})           & ResNet-50                      & \multicolumn{1}{c}{16.4} & 21.1  & 19.4  & 17.8  & 15.5  & 13.4  \\ 
\textsc{DuMoGa} (\textit{FULL})              & ResNet-50                   & \multicolumn{1}{c}{\textbf{24.4}} & \textbf{31.8}  & \textbf{29.7}  & \textbf{26.8}  & \textbf{23.1}  & \textbf{19.3}  \\ \bottomrule
\end{tabular}

\end{table}

\begin{figure}[!t]
  \centering  \includegraphics[width=0.98\linewidth]{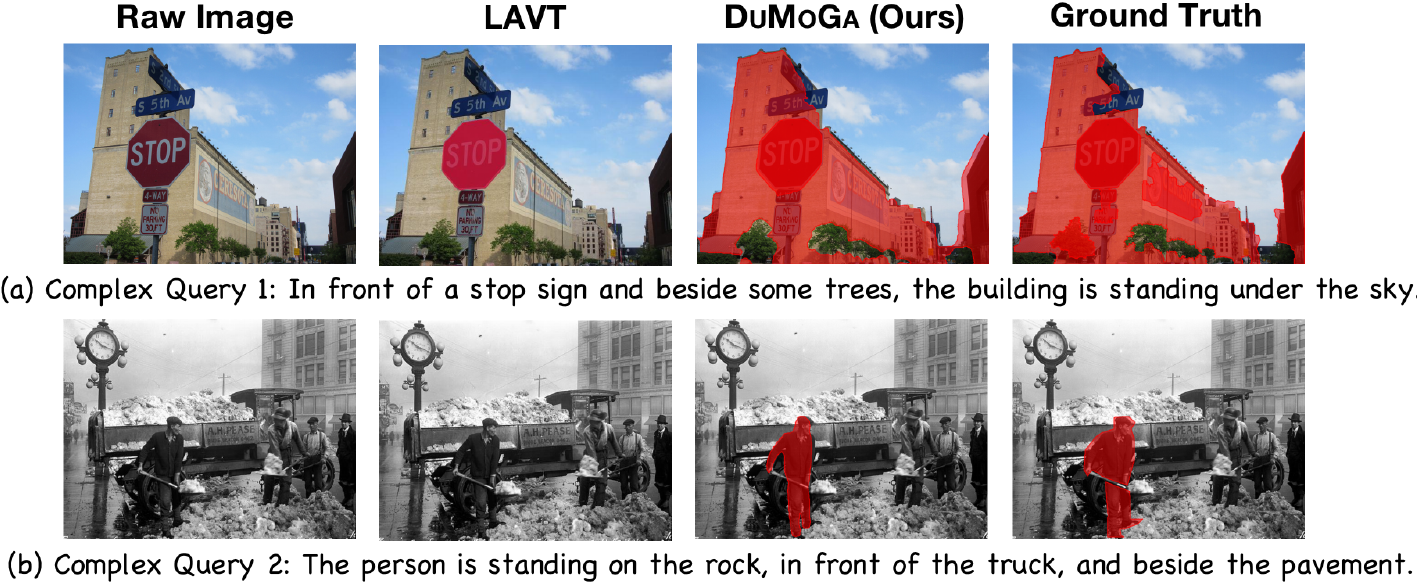}

  \caption{Visualization result of our proposed \textsc{DuMoGa} model on RIS-CQ dataset. It is noteworthy that invalid prediction for LAVT was observed in the case of sample (b).} 
  \label{fig:visualization}

\end{figure}

\subsection{Evaluation Metrics}

For the Referring Image Segmentation task, we follow previous works \cite{liu2017recurrent,margffoy2018dynamic} and adopt Precision@$X$ and IoU to verify the effectiveness.
The IoU calculates intersection regions over union regions of the predicted segmentation mask and the ground truth.
The Precision@$X$ measures the percentage of test images with an IoU score higher than the threshold $X\in\{0.3, 0.4, 0.5, 0.6, 0.7\}$, which focuses on the location ability of the method.

\subsection{Performance Comparison}

Table \ref{tab:overall_performance} shows the performance comparison with other SoTA methods on the RIS-CQ dataset, which includes the following models:

\textbf{LAVT} \cite{yang2022lavt} is a Transformer-based framework for referring image segmentation. Unlike traditional approaches that fuse cross-modal information after feature extraction, LAVT incorporates language-aware visual encoding directly into the model architecture;

\textbf{VPD} \cite{zhao2023unleashing} proposed a new framework that exploits the semantic information of a pre-trained text-to-image diffusion model in visual perception tasks;

\textbf{UNINEXT} \cite{UNINEXT} reformulates diverse instance perception tasks into a unified object discovery and retrieval paradigm and can flexibly perceive different types of objects by simply changing the input prompts;

\textbf{MAttNet} \cite{yu2018mattnet} is a two-stage method that first extracts multiple instances by using an instance segmentation network Mask RCNN \cite{he2017mask}, then utilizes language features to select the target from the extracted instances.
Due to the space limitation, we only list selected SoTA methods, more details can be found in \textbf{Supplementary Materials}.

As shown in Table  \ref{tab:overall_performance}, our method achieves remarkably superior performance on RIS-CQ dataset compared with other SoTA methods. 
Also our \textsc{DuMoGa} model is based on ResNet-50 \cite{he2016deep}, which is not so strong as Swin-B \cite{liu2021swin}, \textsc{DuMoGa} achieves over 100\% to 200\% performance gain in all metrics compared with LAVT \cite{ding2021vision} on RIS-CQ dataset. With stronger backbone models, the performance of our \textsc{DuMoGa} model can be further improved.

We also provide the visualization of the output of our \textsc{DuMoGa} model in Figure \ref{fig:visualization}. Compared with LAVT \cite{ding2021vision}, our \textsc{DuMoGa} can accurately locate the referring object based on the comprehensive understanding of visual-textual data, rather than response to partial words in the query (such as the \textit{stop sign} in Figure \ref{fig:visualization} (a)).

\subsection{In-depth Analysis}



\textbf{Q1: How to control the annotation quality?} 
The annotation process for the RIS-CQ dataset spanned over a period of 6 months and involved the collaboration of 5 undergraduate students. To ensure the production of high-quality annotations, the annotators were supervised and adhered to specific principles. Firstly, all annotators underwent rigorous training before commencing the actual annotation process. Secondly, separate annotators were assigned for query and object selection annotations. Annotators responsible for object selection were instructed to review the quality of queries, ensuring that unclear or poor queries were either rectified or eliminated. This approach simulated the evaluation process, guaranteeing the reasonableness of the queries while avoiding subjective annotations. Lastly, guidelines were provided for the maximal lengths of language queries, specifying limits of over 20 words for queries and over 5 for objects/background labels.
We provide more details in the \textbf{Supplementary Materials}.

\textbf{Q2: What is the necessity of complex query?} \textbf{On the one hand}, current RIS models are sensitive to language queries. Even input two different queries which refer to the same object but contain information in different granularities, current RIS models will output different regions. So we need to construct a new RIS dataset with complex queries to validate the robustness of the proposed RIS models. 
\textbf{On the other hand}, it is not the final path to general RIS model if restricted to train models with downstream annotated benchmark datasets, such as RefCOCO with limited object classes and short query length. It's time to make a further step to real applications by combining with large pre-trained models and propose a novel and informative RIS dataset with more relations among objects.

\section{What To Do Next?}

With the emergence of large pre-trained models (such as GPT-4~\cite{bubeck2023gpt4} and SAM~\cite{kirillov2023segment}), the semantic understanding ability of dealing with multi-modal data has been rapidly enhanced. 
Then, utilizing large pre-trained models is an irresistible trend due to their superiority in open-vocabulary scenarios. In the next step, we can develop lightweight RIS modules (based on the Graph Alignment and Feature Alignment modules in our \textsc{DuMoGa} model) to adapt large pre-trained models in a plug-and-play manner.
Besides, on the basis of our proposed RIS-CQ dataset, we can explore more scenarios in real applications, such as referring object understanding in video and audio modality. And evaluate the robustness of  proposed RIS models even with none or multiple referring objects according to the complex queries.

\section{Related Work}

Referring image segmentation aims to segment a target region (\eg object or stuff) in an image by understanding a given natural linguistic expression, which was first introduced by \cite{hu2016segmentation}.
Early works \cite{liu2017recurrent,li2018referring,margffoy2018dynamic} first extracted visual and linguistic features by CNN and LSTM, respectively, and directly concatenated two modalities to obtain final segmentation results by an FCN \cite{long2015fully}.
In MAttNet \cite{yu2018mattnet}, Yu \etal proposed a two-stage method that first extracts instances using Mask R-CNN \cite{he2017mask}, and then adopts linguistic features to choose the target from those instances.
Then, a series of works are proposed to adopt the attention mechanism.
EFNet \cite{feng2021encoder} designs a co-attention mechanism to use language to refine the multi-modal features progressively,
which can promote the consistency of the cross-modal information representation.
Some recent works leverage transformer \cite{vaswani2017attention} to deal with the RES task with satisfying performance. VLT \cite{ding2021vision} employs a transformer to build a network with an encoder-decoder attention mechanism for enhancing the global context information.
All these methods are trained with RIS datasets with simple queries.

\section{Conclusion}

In this paper, we propose a novel benchmark dataset for Referring Image Segmentation with Complex Queries (RIS-CQ).
The RIS-CQ dataset is of high quality and large scale, which challenges the existing RIS with complex, specific and informative language queries, and enables a more realistic scenario of RIS research.
Besides, we propose a novel SoTA framework to task the RIS-CQ dataset, called dual-modality graph alignment model (\textsc{DuMoGa}), which effectively captures the intrinsic correspondence of the semantics of the two modalities.
Experimental results and analyses demonstrate the necessity of our proposed dataset and the model.

{\small
\bibliographystyle{plainnat}
\bibliography{main}
}

\appendix
\clearpage

\section{Performance Comparison}

Table \ref{tab:overall_performance} shows the performance comparison with other SoTA methods on the RIS-CQ dataset, which includes the following models:

\textbf{LAVT} \cite{yang2022lavt} is a Transformer-based framework for referring image segmentation. Unlike traditional approaches that fuse cross-modal information after feature extraction, LAVT incorporates language-aware visual encoding directly into the model architecture;

\textbf{VPD} \cite{zhao2023unleashing} proposed a new framework that exploits the semantic information of a pre-trained text-to-image diffusion model in visual perception tasks;

\textbf{UNINEXT} \cite{UNINEXT} reformulates diverse instance perception tasks into a unified object discovery and retrieval paradigm and can flexibly perceive different types of objects by simply changing the input prompts;

\textbf{MAttNet} \cite{yu2018mattnet} is a two-stage method that first extracts multiple instances by using an instance segmentation network Mask RCNN \cite{he2017mask}, then utilizes language features to select the target from the extracted instances.








As shown in Table  \ref{tab:overall_performance}, our method achieves remarkably superior performance on CIU-CQ dataset compared with other SoTA methods. 
Also our \textsc{DuMoGa} model is based on ResNet-50 \cite{he2016deep} and ResNet-101, which are not as strong as Swin-B \cite{liu2021swin}, \textsc{DuMoGa} achieves over 100\% to 200\% performance gain in all metrics compared with LAVT \cite{ding2021vision} on RIS-CQ dataset. With stronger backbone models, the performance of our \textsc{DuMoGa} model can be further improved.

\begin{table}[!b]
\centering
\fontsize{9}{12}\selectfont
 \setlength{\tabcolsep}{2.5mm}

\caption{Comparison with state-of-the-art methods in terms of overall IoU on three benchmark datasets. \textit{GA} represents Graph Alignment, \textit{FA} represents Feature Alignment, and \textit{Full} represents Graph Alignment + Feature Alignment.}
\label{tab:overall_performance}

\begin{tabular}{l|c|cccccc}
\toprule
\multirow{2}{*}{Method} & \multirow{2}{*}{Backbone Model} & \multicolumn{6}{c}{Refer Image Understanding}                     \\ \cline{3-8} 
                        &                                 & \multicolumn{1}{c|}{mIoU} & P@0.3 & P@0.4 & P@0.5 & P@0.6 & P@0.7 \\ \hline
MAttNet                & ResNet-101                    & \multicolumn{1}{c|}{8.00} & 9.61  & 7.90  & 6.15  & 5.51  & 4.58  \\ \hline
LAVT                  & SWIN-B                        & \multicolumn{1}{c|}{21.2} & 26.6  & 21.6  & 17.1  & 13.7   & 10.9     \\ \hline

VPD                   & U-Net                      & \multicolumn{1}{c|}{24.0} & 29.5  & 27.5  & 23.8  & 21.5   & 19.3  \\ \hline

UNINEXT                 & ResNet-50                      & \multicolumn{1}{c|}{19.8} & 22.3  & 21.8  & 21.0  & 19.9   & 19.2   \\ \hline

DuMoGa (\textit{GA})             & ResNet-50                       & \multicolumn{1}{c|}{15.0} & 19.5  & 18.3  & 16.7  & 14.4  & 11.7  \\ \hline
DuMoGa (\textit{FA})            & ResNet-50                       & \multicolumn{1}{c|}{16.4} & 21.1  & 19.4  & 17.8  & 15.5  & 13.4  \\ \hline
DuMoGa (\textit{FULL})              & ResNet-50                       & \multicolumn{1}{c|}{24.4} & 31.8  & 29.7  & 26.8  & 23.1  & 19.3  \\ \hline
DuMoGa  (\textit{GA})            & ResNet-101                      & \multicolumn{1}{c|}{16.8}     & 21.8      & 21.2      & 19.0      & 16.3      & 13.4      \\ \hline
DuMoGa  (\textit{FA})          & ResNet-101                      & \multicolumn{1}{c|}{17.5}     & 21.6      & 20.3      & 18.7      & 16.9      & 14.7      \\ \hline
DuMoGa  (\textit{FULL})             & ResNet-101                      & \multicolumn{1}{c|}{\textbf{25.3}}     & \textbf{32.4}      & \textbf{30.4}      & \textbf{28.0}      & \textbf{25.2}      & \textbf{20.5}      \\ \bottomrule
\end{tabular}

\end{table}

The DuMoGa framework is meticulously designed with a plug-and-play approach, aiming to address the process of efficient dual-modality alignment. The framework offers flexibility by allowing the substitution of the VCTree \cite{tang2019learning} method, employed for scene graph generation, and the visual backbone, which is responsible for extracting visual feature representations, with alternative methodologies and backbone architectures. This adaptability makes the DuMoGa framework compatible with a wide range of approaches, thereby accommodating the evolving landscape of research in the field. By leveraging different backbone models, as demonstrated in Table 1, the DuMoGa framework exhibits its versatility. Through the utilization of more sophisticated feature extraction techniques, such as employing ResNet-101 instead of ResNet-50, the DuMoGa framework achieves superior performance across various evaluation metrics on the RIU dataset. This empirical evidence underscores the efficacy of incorporating more fine-grained feature extraction methods within the DuMoGa framework.

In essence, our DuMoGa framework exhibits excellent adaptability in an era characterized by the emergence of diverse feature extraction methods. For instance, the SAM-series of works \cite{kirillov2023segment} have showcased remarkable prowess in instance segmentation tasks, yielding exceptional performance. Leveraging the advantages of the SAM models, they can serve as robust backbone models for the DuMoGa framework. This integration not only enhances the quality of generated scene graphs but also improves the extraction of visual features, leading to more accurate and comprehensive inference results. The integration of SAM into the DuMoGa framework represents a significant advancement in the field, showcasing the potential for synergistic collaboration between different methods. It demonstrates how the adaptability and extensibility of the DuMoGa framework facilitate the integration of state-of-the-art techniques, ultimately leading to improved performance and promising avenues for further research in the community.

\textbf{Effectiveness of each component in our \textsc{DuMoGa} method.} 
As shown in Table \ref{tab:overall_performance}, we list three variants of our \textsc{DuMoGa} model. \textsc{DuMoGa} (\textit{GA}), \textsc{DuMoGa} (\textit{FA}), and \textsc{DuMoGa} (\textit{Full}) represent \textsc{DuMoGa} with only graph alignment, \textsc{DuMoGa} with only feature alignment, and \textsc{DuMoGa} with graph alignment + feature alignment, respectively. To evaluate the effectiveness of graph alignment process, we train a \textsc{DuMoGa} (\textit{GA}) only considers the information contained in scene graphs and semantic tree graphs, which only have a few MLPs to complete the task. The same evaluation process for our feature alignment process, we only use the visual feature extracted from the SGG model and the sentence embedding from BERT. These two simple models still outperform traditional RIS methods by a great margin. The reasons is that as the complexity of query sentence increasing, traditional methods may be mislead by the sophisticated logic and finally get low performances, however, the parsed dependency tree graph is likely to shares higher similarity with the generated scene graph for its richer information within. As a result, our method easily achieves better performance.

\section{Referring Query Generation}
\begin{figure}[t]
  \centering  \includegraphics[width=\linewidth]{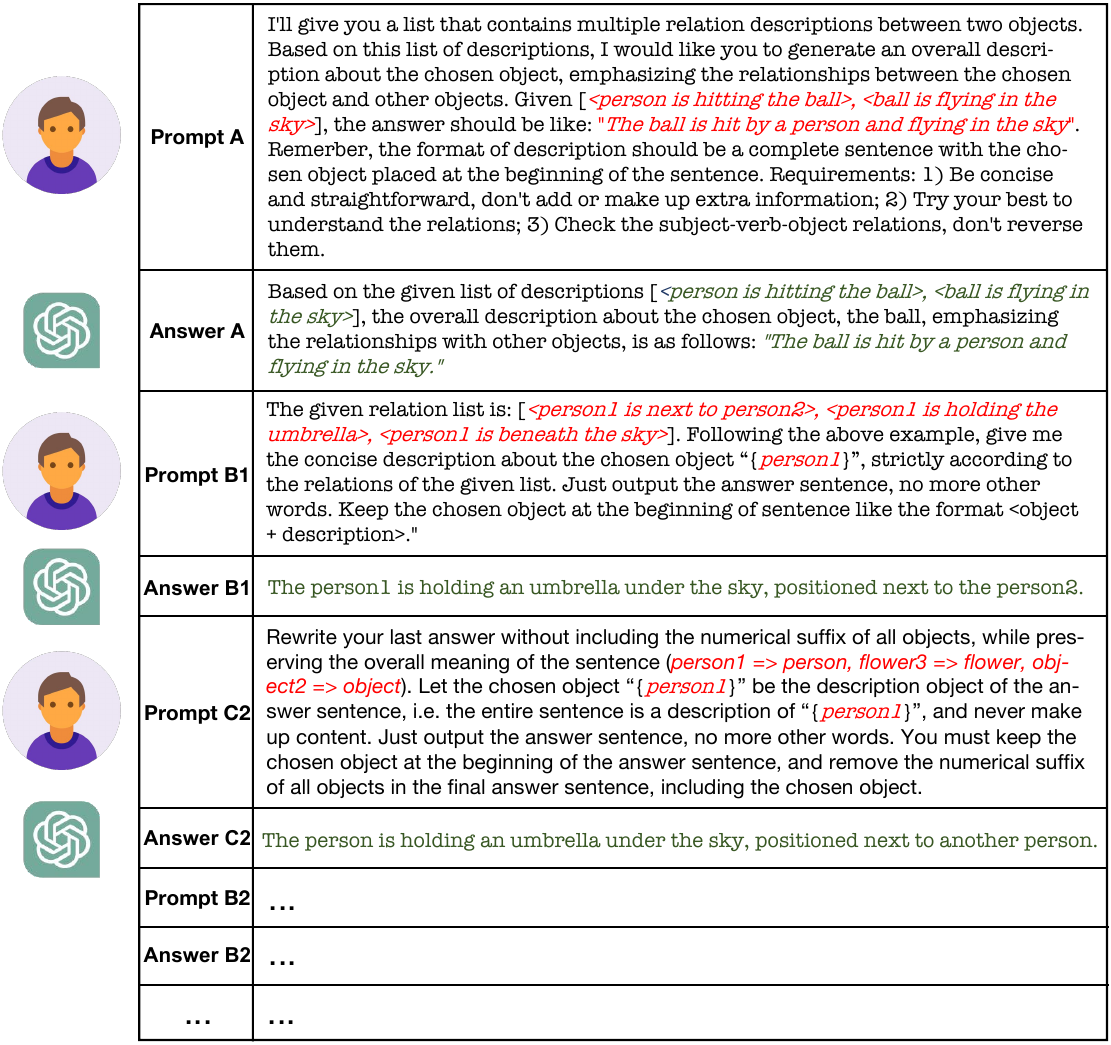}

  \caption{An example of generating descriptions for a specified object based on the relations between objects provided by a scene graph model using ChatGPT.} 
  \label{fig:prompt}

\end{figure}

Figure \ref{fig:prompt} depicts the procedural flow of generating referring queries using the outputs of a scene graph model. \textbf{Initially}, we engage in a dialogue with ChatGPT using Prompt A as the opening prompt. Prompt A serves the purpose of clarifying the task we seek ChatGPT's assistance with and outlining the associated requirements. Additionally, examples are provided to harness the powerful in-context learning capabilities of ChatGPT, aiming to achieve reasoned inferences that align with our specified needs. \textbf{Subsequently}, we instruct ChatGPT, following Prompt A, regarding the object we intend to describe and its relevant relations to other objects. This enables ChatGPT to generate descriptions of the designated objects as per the instructions outlined in the prompt. It is noteworthy that, to alleviate the inferential burden on ChatGPT, the triplets generated by the scene graph model (e.g., <person, book, holding>) are directly transformed into complete sentences (e.g., "the person is holding the book"). Moreover, to differentiate between objects of the same category, numerical suffixes such as "person2" or "cow3" are assigned. This facilitates relatively precise object descriptions leveraging relations, thereby significantly reducing the need for labor-intensive manual annotation of specific object attributes. \textbf{Finally}, we perform a rewrite on the initial answer generated by ChatGPT in Answer B, removing the numeric suffix corresponding to the object (e.g., "<person1 is next to person2" becomes "the person is next to another person," and "<the floor has person1, person2, and person3 standing on it" becomes "the floor has multiple people standing on it").

\section{Potential Impact and Limitations}

\textbf{Potential Impact.} 
Referring Image Segmentation can track people in the scene and understand their relations with the surrounding
objects via language description, it has the potential to be used to violate people’s privacy. The possible application to the surveillance system can also bring negative societal impacts.

\textbf{Limitation.} In this paper, we explore referring image segmentation with complex queries for only one object. However, referring multiple objects with complex language queries is also an important task in real applications, which we will leave as future work.

\newpage

\end{document}